\documentclass[runningheads]{llncs}
\usepackage{graphicx}
\usepackage{amsmath,amssymb}
\usepackage{multirow}
\begin{document}
\title{Focal and Global Spatial-Temporal Transformer \\for Skeleton-based Action Recognition}
\titlerunning{FG-STFormer for Skeleton-based Action Recognition}
%
\author{Zhimin Gao\inst{1}\and
Peitao Wang\inst{1}\and
Pei Lv\inst{1}\and
Xiaoheng Jiang\inst{1}\and
Qidong Liu\inst{1}\and
Pichao Wang\inst{2}\and
Mingliang Xu\inst{1}\thanks{Corresponding author}\and
Wanqing Li\inst{3}
}
\authorrunning{Z. Gao et al.}
%
\institute{Zhengzhou University, Zhengzhou, China\\\email{zhimingao113@gmail.com}, \email{wptao\_98@163.com},\\\email{\{ielvpei,jiangxiaoheng,ieqdliu,iexumingliang\}@zzu.edu.cn} \and
DAMO Academy, Alibaba Group (U.S.) Inc\\
\email{pichaowang@gmail.com}\\\and
AMRL, University of Wollongong,
Wollongong, Australia\\
\email{wanqing@uow.edu.au}}
\maketitle              
\begin{abstract}
Despite great progress achieved by transformer in various vision tasks, it is still underexplored for skeleton-based action recognition with only a few attempts. Besides, these methods directly calculate the pair-wise global self-attention equally for all the joints in both the spatial and temporal dimensions, undervaluing the effect of discriminative local joints and the short-range temporal dynamics. In this work, we propose a novel \textbf{F}ocal and \textbf{G}lobal \textbf{S}patial-\textbf{T}emporal Trans\textbf{former} network (FG-STFormer), that is equipped with two key components: (1) FG-SFormer: focal joints and global parts coupling spatial transformer. It forces the network to focus on modelling correlations for both the learned discriminative spatial joints and human body parts respectively. The selective focal joints eliminate the negative effect of non-informative ones during accumulating the correlations. Meanwhile, the interactions between the focal joints and body parts are incorporated to enhance the spatial dependencies via mutual cross-attention. (2) FG-TFormer: focal and global temporal transformer. Dilated temporal convolution is integrated into the global self-attention mechanism to explicitly capture the local temporal motion patterns of joints or body parts, which is found to be vital important to make temporal transformer work. Extensive experimental results on three benchmarks, namely NTU-60, NTU-120 and NW-UCLA, show our FG-STFormer surpasses all existing transformer-based methods, and compares favourably with state-of-the-art GCN-based methods.

\end{abstract}
\section{Introduction}
Human action recognition has long been a crucial and active research field in video understanding since it has a broad range of applications, such as human-computer interaction, intelligent video surveillance and robotics~\cite{poppe2010survey,carreira2017quo,wang2018rgb}. In recent years, skeleton-based action recognition has gained increasing attention with advent of cost-effective depth cameras like Microsoft Kinect~\cite{zhang2012microsoft} and advanced pose estimation techniques~\cite{cao2019openpose}, which make skeleton data more accurate and accessible. By representing the action as a sequence of joint coordinates of human body, the highly abstracted skeleton data is compact and robust to illumination, human appearance changes and background noises.

Effectively modelling the spatial-temporal correlations and dynamics of joints is crucial for recognizing actions from skeleton sequences. The dominant solutions to it in recent years are the graph convolutional networks (GCNs)~\cite{yan2018spatial}, as they can model the irregular topology of the human skeleton. Via designing advanced graph topology or traversal rules, the recognition performance is greatly improved by GCN-based methods~\cite{liu2020disentangling,song2021constructing}. 
Meanwhile, the recent success of Transformer~\cite{vaswani2017attention} has gained significant interest and performance boost in various computer vision tasks~\cite{dosovitskiy2020image,liu2021swin,carion2020end,neimark2021video}. For skeleton-based action recognition, one would expect that the self-attention mechanism in transformer shall naturally capture effective correlations of joints in both spatial and temporal dimensions for action categorization, without enforcing the articulating constrains of human body like GCN. However, there are only a few transformer-based attempts~\cite{shi2020decoupled,plizzari2021skeleton,zhang2021stst}, and they devise hybrid model of GCN and transformer~\cite{plizzari2021skeleton} or multi-task learning framework~\cite{zhang2021stst}. How to utilize self-attention to learn effective spatial-temporal relations of joints and representative motion features is still a thorny problem. Moreover, most of these Transformer based methods directly calculate the global one-to-one relations of joints for spatial and temporal dimensions respectively. Such strategy undervalues the spatial interactions of discriminative local joints and short-term temporal dynamics for identifying crucial action-related patterns. On the one hand, since not all joints are informative for recognizing actions~\cite{liu2017global,jiang2015informative}, these methods suffer from the influence of irrelevant or noisy joints by accumulating the correlations with them via attention mechanism, which could harm the recognition. On the other hand, with the fact that the vanilla transformer lacks of inductive bias~\cite{liu2021swin} to capture the locality of temporal structural data, it is difficult for these methods to directly model effective temporal relations of joints globally over long input sequence.

\begin{figure}[hbt]
\centering
\includegraphics[scale=0.46]{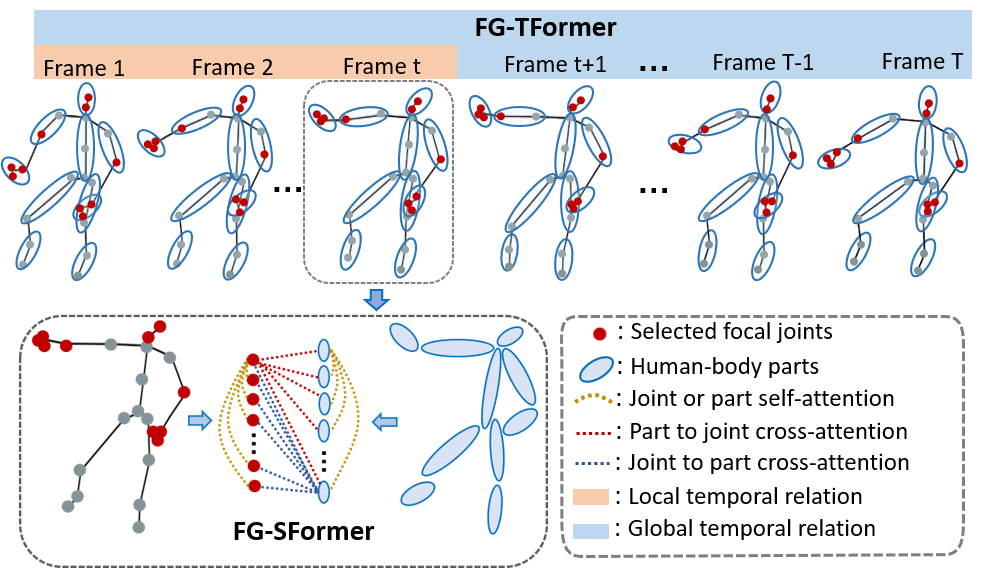}
\caption{The proposed FG-SFormer (bottom) learns correlations for both adaptively selected focal joints and body parts, as well as the joint-part interactions via cross-attention. FG-TFormer (top) models the explicit local temporal relations of joints or parts, as well as the global temporal dynamics.}
\label{fig:idea}
\end{figure}

To tackle these issues, we propose a novel end-to-end \textbf{F}ocal and \textbf{G}lobal \textbf{S}patial-\textbf{T}emporal Trans\textbf{former} network, dubbed as FG-STFormer, to effectively capture relations of the crucial local joints and the global contextual information in both spatial and temporal dimensions for skeleton-based action recognition. It is composed of two components: FG-SFormer and FG-TFormer. Intuitively, each action can be distinguished by the co-movement of: (1) some critical local joints, (2) global body parts, and (or) (3) joint-part interaction. For example, as shown in Fig.~\ref{fig:idea}, actions such as {\it taking a selfie} and {\it kicking} mainly involve important joints of hands, head and feet, as well as related body parts of arms and legs, while the actions like {\it sit down} primarily require understanding of body parts cooperation and dynamics. Based on the above observations, at the late stage of the network, we adaptively sample the informative spatial local joints (focal joints) for each action, and force the network to focus on modelling the correlations among them via multi-head self-attention without involving non-informative joints. Meanwhile, in order to compensate for the missing global co-movement and spatial structure information, we incorporate the dependencies among human body parts using self-attention. Furthermore, interactions between the body parts and the focal joints are explicitly modelled via mutual cross-attention to enhance their spatial collaboration. All of these are achieved by the proposed FG-SFormer. 

The FG-TFormer is designed to model the temporal dynamics of joints or body parts. It is found that straightforwardly using the vanilla temporal transformer leads to ineffective temporal relations and poor recognition performance. We found one of the key culprits lying in the absence of local bias, making it challenging for transformer to focus on effective temporal motion patterns in the long input. Taking these factors into consideration, we integrate the dilated temporal convolutions into multi-head self-attention mechanism to explicitly encode the short-term temporal motions of a joint or part from their neighbors respectively, which equips transformer with local inductive bias. The short-range feature representations of all the frames are further fused by the global self-attention weights to embrace the global contextual motion information into the representations. Thus, the designed strategy enables transformer to learn both important local and effective global temporal relations of the joints and human body parts in a unified structure, which is validated critical to make temporal transformer work. 


To summarize, the contributions of this work lie in four aspects:
\begin{enumerate}
    \item We propose a novel FG-STFormer network for skeleton-based action recognition, that can effectively capture the discriminative correlations of focal joints as well as the global contextual motion information in both the spatial and temporal dimensions. 
    \item We design a focal joints and global parts coupling spatial transformer, namely FG-SFormer, to model the correlations of adaptively selected focal joints and that of human body parts. The joint-part mutual cross-attention is integrated to enhance the spatial collaboration. 
    \item We introduce a FG-TFormer to explicitly capture both the short and long range temporal dependencies of the joints and body parts effectively.
    \item The extensive experimental results on three datasets highlight the effectiveness of our method, that surpasses all existing transformer-based methods.
\end{enumerate}
\section{Related Work}
\label{sec:related-work}
\textbf{Skeleton-based Action Recognition.} With great progress achieved in skeleton-based action recognition, existing works can be broadly divided into three groups, i.e., RNNs, CNNs, and GCNs based methods. RNNs concatenate the coordinates of all joints in one frame and treat the sequence as time series~\cite{du2015hierarchical,zhu2016co,lee2017ensemble,zhang2017view,li2018independently}. Some works design specialized network structure, like trees~\cite{liu2016spatio,wang2017modeling} to make RNN aware of spatial information.
CNN based methods transform one skeleton sequence to a pseudo-image via hand-crafted manners~\cite{wang2016action,ke2017new,li2017skeleton,liu2017enhanced,hou2016skeleton,duan2021revisiting,li2021memory}, and then use popular networks to learn spatial and temporal dynamics in it. 

The appearance of GCN based methods, like ST-GCN~\cite{yan2018spatial}, enables more natural spatial topology representation of skeleton joints by organizing them as a non-Euclidean graph. The spatial correlation is modelled for bone-connected joints. As the fixed graph topology (or adjacency matrix) is not flexible to model the dependencies among spatially disconnected joints, many subsequent methods focus on designing high-order or multi-scale adjacency matrix~\cite{li2019actional,gao2019optimized,liu2020disentangling,huang2020spatio,li2019spatio}, and dynamically adjusted graph topology~\cite{shi2019two,li2019actional,ye2020dynamic,zhang2020semantics,chen2021channel}. 
Nevertheless, these manually devised joint traversal rules limit the flexibility to learn more effective spatial-temporal dynamics of joints for action recognition.

\textbf{Transformer based Methods.} 
Several recent works extend Transformer~\cite{vaswani2017attention} to spatial and temporal dimensions of skeleton-based action recognition. Among them, DSTA~\cite{shi2020decoupled} is the first to use self-attention to learn joint relations, whereas in practice spatial transformer interleaved with temporal convolution is employed for some typical datasets. ST-TR~\cite{plizzari2021skeleton} adopts a hybrid architecture of GCN and transformer in a two-stream network, with each stream replacing the GCN or temporal convolution with spatial or temporal self-attention. STST~\cite{zhang2021stst} introduces a transformer network that the spatial and temporal dimensions are parallelly separated. Besides, the network is trained together with multi-task self-supervised learning tasks. 
\section{Proposed Method}
In this section, we first briefly review the basic spatial and temporal Transformer blocks (referred to as Basic-SFormer and Basic-TFormer blocks respectively) used by most existing skeleton-based action recognition methods~\cite{plizzari2021skeleton,shi2020decoupled}, which is also the basics of our network. Then the proposed Focal and Global Spatial-Temporal Transformer (FG-STFormer) is introduced in detail.

\subsection{Basic Spatial-Temporal Transformer on Skeleton Data} \label{subsec:basicformer}

{\bf Vanilla Transformer (V-Former) Block.}
The vanilla transformer~\cite{vaswani2017attention} block consists of two important modules:  multi-head self-attention (MSA) and point-wise feed-forward network (FFN). Let an input composed of $N$ elements and $C$-dimensional features be $X \in \mathbb{R}^{N \times C}$. For a MSA having $H$ heads, $X$ is first linearly projected to a set of queries $Q$, keys $K$ and values $V$. Then, the scaled dot-product attention of head $h$ is calculated as:
\begin{equation}\label{eqn:sa}
\begin{aligned}
\text {Attention}(Q^h,K^h,V^h) = \text {softmax}(\frac{Q^h {K^h}^T}{\sqrt{d}})V^h = A^hV^h~,    
\end{aligned}
\end{equation}
where $Q^h$, $K^h$, $V^h\in \mathbb{R}^{N \times d}$ with $d = C/H$ being the feature dimension of one head. $A^h \in \mathbb{R}^{N \times N}$ is the attention map.

MSA concatenates the output of all the heads and feeds into FFN module, that generally consists of a number of linear layers to transform the features. 

{\bf Basic Spatial Transformer (Basic-SFormer) Block.} 
For a skeleton sequence of $T$ frames and $N$ joints, let the input of $C$-dimension be $X = \left\{X_t\in \mathbb{R}^{N \times {C}}\right\}_{t=1}^{T} $. The Basic-SFormer block extends the V-Former block~\cite{vaswani2017attention} to spatial dimension. It computes the inter-joint correlations for each frame $X_{t}$ via Eq.~(\ref{eqn:sa}) and generates an attention map $A_t^h \in \mathbb{R}^{N \times N}$, with each element $(A_t^h)_{ij}$ representing the spatial correlation score between joints $i$ and $j$. 
Then, the features of each joint are updated as the weighted sum of values of all the joints. For the entire skeleton sequence, $T$ spatial attention maps are produced. 

{\bf Basic Temporal Transformer (Basic-TFormer) Block.} By extending the V-Former to the temporal dimension, one Basic-TFormer learns global-range dynamics of a joint along the entire sequence. It rearranges the input as $X = \left\{X_n \in \mathbb{R}^{T \times C}\right\}_{n=1}^{N}$ to tackle temporal dimension. With Eq.~(\ref{eqn:sa}), one of the $N$ attention map $A_n^h \in \mathbb{R}^{T \times T}$ is computed for the $n^{\text {th}}$ joint. Each row in it stands for the dependencies of this joint across all the frames. 


\subsection{Focal and Global Spatial-Temporal Transformer Overview}
\begin{figure}[htp]
\centering
\includegraphics[scale=0.36]{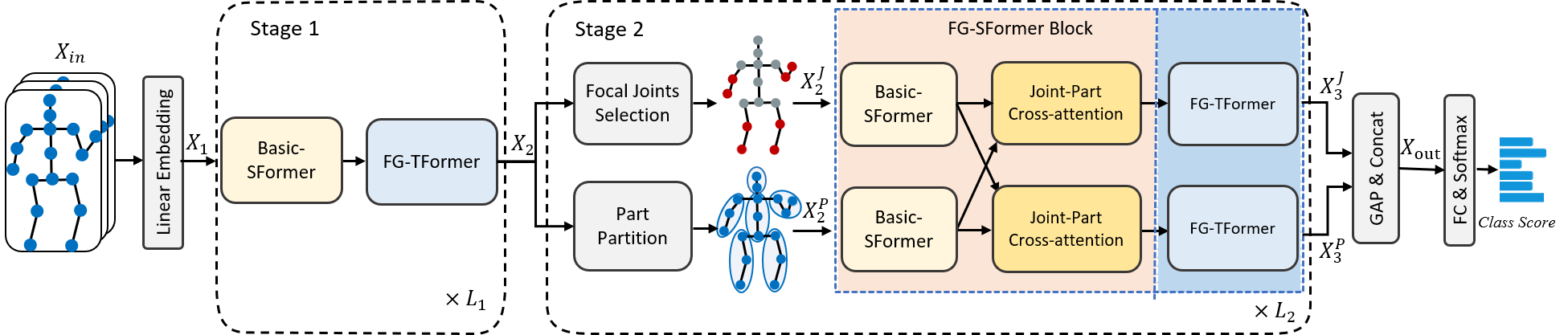}
\caption{Architecture of the proposed Focal and Global Spatial-Temporal Transformer (FG-STFormer). $L_1$ and $L_2$ are the number of layers in Stage 1 and Stage 2 respectively.}
\label{fig:architecture}
\end{figure}

The overview of the proposed FG-STFormer network is depicted in Fig.~\ref{fig:architecture}. It consists of two stages, in which our two primary components are FG-SFormer block and FG-TFormer block. The former is designed for the network late stage to model both the correlations of the sampled focal joints and the co-movement of human body parts globally in spatial dimension, as well as the interactions between the focal joints and body parts. The latter is devised to learn important local relations explicitly and global motion dynamics in temporal dimension, and is used in both stages. Therefore, the two stages are assigned specific responsibilities. That is, stage 1 aims to learn correlations for all joint pairs as generally done, so as to provide effective representations for stage 2 to mine reliable focal joints and part embeddings. Stage 2 targets at modelling both the discriminative relations among focal joints and the global movement information of body parts. These two stages cooperate with each other to make the network learn discriminative and comprehensive spatial-temporal motion patterns for recognition. 

Specifically, given a raw skeleton sequence $X_{\text {in}} \in \mathbb{R}^{N \times T \times {C_0}}$, a linear layer is first applied to project it from the 2D or 3D joint coordinates of $C_0$ to a higher dimension $C_1$, generating feature $X_1 \in \mathbb{R}^{N \times T \times {C_1}}$. Then, $X_1$ goes through the two successive stages of FG-STFormer. Stage 1 sequentially stacks $L_1$ layers with each consisting of a Basic-SFormer block and an our FG-TFormer block. 

At the end of stage 1, we obtain the high-level feature representations $X_2 \in \mathbb{R}^{N \times T \times {C_2}}$. It is then passed into stage 2, where the network is split into two branches. One branch adaptively selects $K$ focal joints for each frame of the sequence and discards the remaining non-informative ones, producing features $X_2^J \in \mathbb{R}^{K \times T \times {C_2}}$. Meanwhile, the other branch partitions the joints into $P$ global-level human body parts and generates feature tensor $X_2^P \in \mathbb{R}^{P \times T \times {C_2}}$. $X_2^J$ and $X_2^P$ are then passed through $L_2$ layers that interleave FG-SFormer and FG-TFormer blocks. In particular, one FG-SFormer block consists of a Basic-SFormer sub-block and a joint-part cross-attention sub-block to sufficiently model the spatial interaction information of actions. Stage 2 then produces output features $X_3^J$ and $X_3^P$ from the two branches respectively. They are applied global average pooling (GAP), and then concatenated along feature channels producing features $X_{\text{out}} \in \mathbb{R}^{1 \times 1 \times {C_{\text {out}}}}$. With which, FG-STFormer finally performs classification using two fully connected layers and a Softmax classier. 

\subsection{Focal and Global Spatial Transformer (FG-SFormer)} 
The proposed FG-SFormer block designed for network stage 2 learns critical and comprehensive spatial structure and motion patterns based on facts in two aspects. For one aspect, there is often a subset of key joints that play a vital role in action categorization~\cite{liu2017global,jiang2015informative}, while the other joints are irrelevant or even noisy for action analysis. Especially, for transformer-based methods, the features of one joint could be influenced by those non-informative ones when integrating features of all the joints. Therefore, it is beneficial to identify the focal joints and concentrate on them at the deep layers of the network after the shallow layers have sufficiently learned the relationships among all the joints.

For the other aspect, it is not enough to just focus on the movement of focal joints. The movement of human body parts carry crucial global contextual motion information for recognizing an action~\cite{du2015hierarchical,huang2020part}. Meanwhile, the interactions between joints and parts convey rich kinematic information, that could be exploited to fully mine action-related patterns. 

Therefore, we propose to learn relations for adaptively identified focal joints and for human body parts, as well as their interactions in spatial dimension. Three modules to achieve this are designed: (i) Focal joints selection; (ii) Global-level part partition encoding; and (iii) Joint-part cross-attention.

{\bf Focal Joints Selection.} In the joint branch of stage 2, we design a ranking based strategy to adaptively sample the focal joints subset for each frame in an action sequence with the input $X_2 \in \mathbb{R}^{N \times T \times {C_2}}$, and discard the non-informative ones. To achieve this, we leverage a trainable projection vector $W_p \in \mathbb{R}^{C_2 \times 1}$ and sigmoid function to predict the informativeness scores $S \in \mathbb{R}^{N \times T}$ for all the joints in individual frame as: 
\begin{equation}\label{eqn:proj}
\begin{aligned}
S = \text{sigmoid}(X_2 W_p/||W_p||)~,        
\end{aligned}
\end{equation}

Each element $S_{ij}$ represents the informativeness score of $i^{\text {th}}$ joint in $j^{\text {th}}$ frame. The larger the score is, the more informative the joint is. We sort the scores of all the joints for each frame and take the features corresponding to the top $K$ joints having the largest scores to form the features $X_2^J \in \mathbb{R}^{K \times T \times {C_2}}$ of the focal joints subset as:
\begin{equation}\label{eqn:topk}
\begin{aligned}
\text{idx} &= \text{sort}(S,K)~,\\
X_2^J &= X_2(\text{idx},:,:)~,           
\end{aligned}
\end{equation}
where idx is the indices of the selected joints with largest scores.
 
\begin{figure}[ht]
\centering
\includegraphics[scale=0.4]{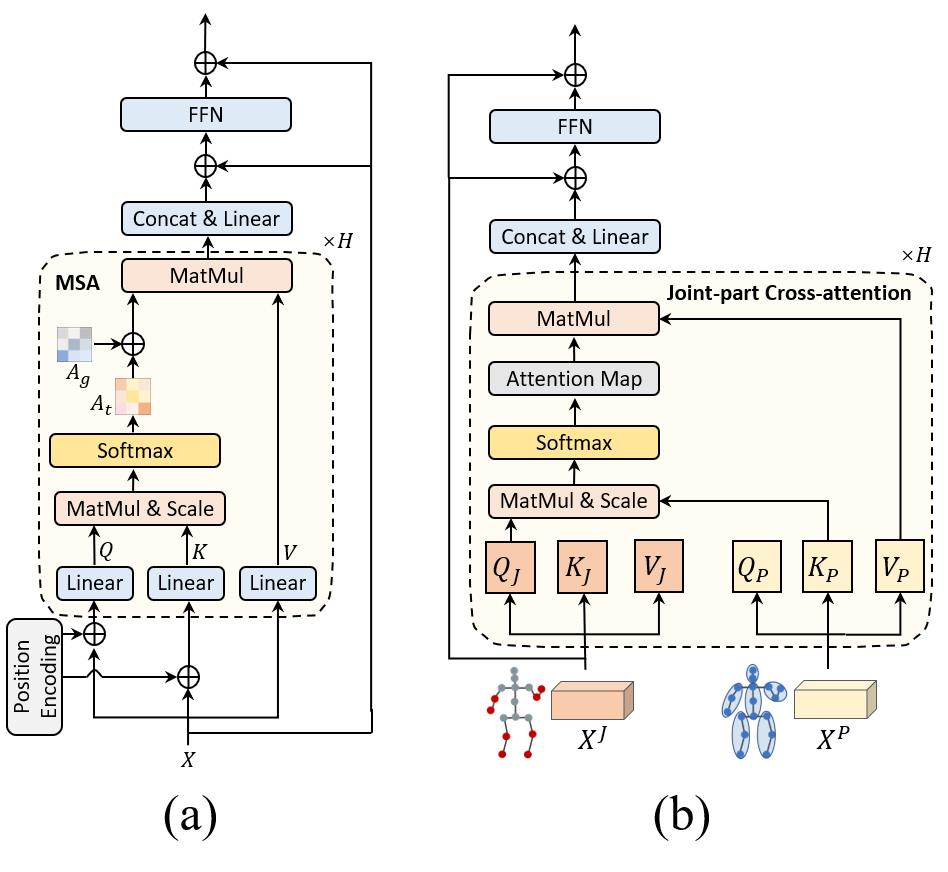}
\caption{(a) The Basic-SFormer block/sub-block used in Stages 1 and 2. (b) The Joint-part cross-attention (JP-CA) sub-block used in FG-SFormer block.}
\label{fig:LGC-SFormer}
\end{figure}

$X_2^J$ is then fed into the Basic-SFormer sub-block introduced in Section~\ref{subsec:basicformer} to calculate the correlations only for those focal joints and update their feature embeddings. The Basic-SFormer block/sub-block used in both stages 1 and 2 is depicted in Fig.~\ref{fig:LGC-SFormer} (a). It uses the sine and cosine position encoding~\cite{vaswani2017attention} to encode the joint 
type information. In the MSA module with $H$ heads, the spatial attention map $A_t$ is calculated for each frame. As in~\cite{shi2020decoupled}, we add a global regularization attention map $A_g$ shared by all the sequences. The FFN module consists of a linear layer followed by applying activation function of Leaky ReLU~\cite{maas2013rectifier}. 

{\bf Global-level Part Partition Encoding.} We explicitly model the correlations between global-level body parts in the other branch of stage 2 in our FG-STFormer. The joints are partitioned into $P$ parts based on the physical skeleton structure and human prior. To obtain feature embeddings of the $P$ parts with $X_2$, we concatenate the features of joints belonging to the same body part and then transform them into one part-level feature embedding via a linear layer shared by all parts. This generates the part embedding $X_2^P \in \mathbb{R}^{P \times T \times {C_2}}$, which is then passed into the Basic-SFormer sub-block shown in Fig.~\ref{fig:LGC-SFormer} (a) to model the one-to-one part relations and update features correspondingly.

{\bf Joint-Part Cross-attention.} To enable information diffusion across the focal joints and body parts to model their co-movement, we devise a joint-part cross-attention sub-block, termed as JP-CA. It uses multi-head cross-attention to interact and diffuse features of the two branches. Here, we present JP-CA from the part branch to the focal joint branch as an example, as shown in Fig.~\ref{fig:LGC-SFormer} (b). For notational convenience, we omit the subscripts of $X_2^J$ and $X_2^P$. Let $Q_J$, $K_J$ and $V_J$ be the queries, keys and values mapped from the joint-branch features $X^J$, and $Q_P$, $K_P$ and $V_P$ be those from the part-branch features $X^P$ respectively. The part-to-joint cross-attention takes the $Q_J$ as queries, and $K_P$ and $V_P$ as keys and values, and is calculated as:
\begin{equation}\label{eqn:jpca}
\begin{aligned}
\text {Attention}(Q_J,K_P,V_P) = \text {softmax}(\frac{{Q_J} {K_P}^T}{\sqrt{d}})V_P = A_{jp}V_P~,    
\end{aligned}
\end{equation}
where $d$ is the feature dimension of one head.

The attention map $A_{jp} \in \mathbb{R}^{K \times P}$ models the joint-part correlations and is used to aggregate part features for each focal joint. Other operations in this sub-block is same as those in the adopted Basic-SFormer sub-block. Notably, JP-CA is adaptive to actions, which is flexible to capture distinct collaborative patterns for input actions. Analogously, the cross-attention from the joint-branch to part-branch can be defined in similar operations. 

\subsection{Focal and Global Temporal Transformer (FG-TFormer)} 
Though temporal transformer has been applied in skeleton-based action recognition in existing works~\cite{zhang2021stst,plizzari2021skeleton,bai2021gcst}, it is rarely effectively deployed solely with spatial transformer in a single-stream architecture or in pure transformer-based models, largely because: (i) it is difficult for the self-attention to directly model effective temporal relations globally for distinguishing actions over the long input sequence; and (ii) the lack of inductive biases of transformer.

\begin{figure}[!hbtp]
\centering
\includegraphics[scale=0.4]{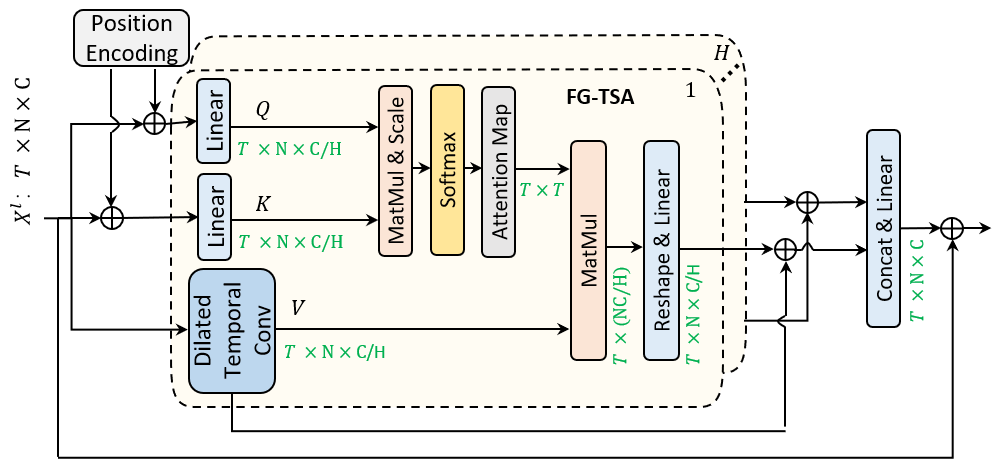}
\caption{The pipeline of the proposed FG-TFormer block.}
\label{fig:lgc-tformer}
\end{figure}

To address these issues, we propose to assist transformer in focusing on both the important local and the global temporal relations of joints explicitly, and design the component of focal and global temporal self-attention (FG-TSA), as depicted in Fig.~\ref{fig:lgc-tformer}. 
It utilizes the dilated temporal convolution to generate the values $V$ in MSA, that works in two aspects: (i) explicitly learning the short-term temporal motion representation of a joint from its neighboring frames; and (ii) introducing beneficial local inductive biases to transformer. Meanwhile, the attention map generated by MSA models the global temporal correlations.
Therefore, the resulting fused joint representations integrate both local temporal relation and global contextual information. The same effect is also achieved for the body part representations when FG-TFormer is applied to the part-branch.

Specifically, let the input feature tensor of the FG-TFormer block in layer $l~(l = 1,2,...,L)$ be $X^l \in \mathbb{R}^{N \times T \times {C}}$, where $L$ is the total number of layers. First, the absolute position encoding is used to encode the temporal order information. And then, for the queries and keys $Q^l_h$, $K^l_h \in \mathbb{R}^{N \times T \times \frac{C}{H}}$ in head $h$ among the $H$ ones, they are generated via the usual linear projection in FG-TSA. Whereas for the $V^l_h$, different from existing works, we utilize dilated temporal convolution with kernel size $k_t \times 1$ and dilation rate $d_t$ to obtain it, denoted as $V^l_h = \text{TCN}_{\text{dilate}}(X^l)_h \in \mathbb{R}^{N \times T \times \frac{C}{H}}$. 
Then, the global self-attention map is calculated via Eq.~(\ref{eqn:sa}) and utilized to fuse the local feature representation $V^l_{h}$. Hence, each joint representation is injected with its global contextual dynamics. 
The features are then reshaped and linearly transformed with $W_h \in \mathbb{R}^{\frac{C}{H}\times \frac{C}{H}}$. This is followed by concatenating the output features of all the heads and conducting linear transform with $W_O \in \mathbb{R}^{{C}\times {C}}$ using activation function of Leaky ReLU. The output of the FG-TFormer block $X^{l+1}$ is obtained by adding the shortcut from the input $X^l$. The whole process is formulated as:
\begin{equation}\label{eqn:lwsa}
\begin{aligned}
&X^{l+1} = \text {Concat}[\text{head}(X^l)_1,...,\text{head}(X^l)_H]W_O + X^l~, \\
\text{head}(X^l)_h &= [\text {Attention}(Q^l_h,K^l_h)\text{TCN}_{\text{dilate}}(X^l)_h]W_h + \text{TCN}_{\text{dilate}}(X^l)_h~.
\end{aligned}
\end{equation}

Besides, we halve the temporal resolution of a sequence during generating $V$ with convolution of stride $2$ when the feature channels are doubled for a FG-TFormer block. This hierarchical structure reduces the computational cost.

\section{Experiments}
\subsection{Dataset}
{\bf NTU-RGB+D 60 (NTU-60)} ~\cite{shahroudy2016ntu} contains 56,880 sequences in 60 classes. It is collected from 40 subjects and provides the 3D locations of 25 human body joints. It recommends two benchmarks for evaluation: (1) Cross-subject (X-Sub): training data is from 20 subjects and test data from the other 20 subjects. (2) Cross-view (X-View): sequences captured by camera views 2 and 3 are taken as training data, and those captured by camera view 1 as testing data. \\
{\bf NTU-RGB+D 120 (NTU-120)}~\cite{liu2019ntu} has 120 classes and 113,945 samples captured from 32 camera setups and 106 subjects. It recommends two benchmarks: (1) Cross-subject (X-Sub): training data is from 53 subjects, and test data from the other 53 subjects. (2) Cross-setup (X-Set): samples with even setup IDs are used for training data, and samples with odd setup IDs for test.\\
{\bf Northwestern-UCLA (NW-UCLA)}~\cite{wang2014cross} is captured by three Kinect cameras from three viewpoins. It contains 1,494 sequences in 10 action categories, with each performed by 10 actors. The same evaluation protocol in~\cite{liu2016spatio} is used: training data from the first two cameras and test data from the other camera. 

\subsection{Implementation Details}
Our FG-STFormer model consists of 8 layers in two stages. Stage 1 contains $L_1 = 6$ layers and stage 2 consists of $L_2 = 2$ layers. The channel dimensions of each layer are 64, 64, 128, 128, 256, 256, 256 and 256. The number of frames is halved at the third and fifth layers. The number of spatial attention heads for the Basic-SFormer and FG-SFormer blocks is set to be 3. Each FG-TFormer block uses 2 attention heads, which adopt temporal kernel size of $k_t = 7$, and dilation rates of $d_t = 1$ and $d_t = 2$ respectively. The numbers of focal joints and body parts in the two branches of stage 2 are $K = 15$ and $P = 10$ respectively. 

All experiments are conducted on one RTX 3090 GPU
with PyTorch framework. We use SGD with Nesterov momentum of 0.9 and weight decay of 0.0005 to train our model for 80 epochs. Warm up strategy is used for the first 5 epochs. The initial learning rate is set to 0.01 and decays by a factor of 10 at the 50th and 70th epochs. For NTU-60 and NTU-120, the batch size is 32. The sequences are sampled to 128 frames, and we use the data pre-processing method in~\cite{chen2021channel}. For NW-UCLA, the batch size is 32, we use the same data pre-processing  in~\cite{wang2014cross}.

\subsection{Ablation Study}
In this section, the effectiveness of individual component of FG-STFormer is evaluated under X-Sub protocol of NTU-60 dataset, using only the joint stream.

{\bf Effectiveness of FG-SFormer Block.} To examine the effectiveness of the proposed FG-SFormer, we evaluate the important components in it, i.e., focal joints selection, part branch and joint-part cross-attention (JP-CA). We employ the Basic-SFormer as baseline, which calculates the correlations for all the joints at every layer of the network without using part branch and JP-CA. For temporal modelling, our FG-TFormer is used. We gradually replace the baseline by adding our designs one-by-one. The experimental results are shown in Table~\ref{tab:LGC-SFormer}. 
\begin{table}[!htbp]
\caption{Ablation study of different components in FG-SFormer block.}
\label{tab:LGC-SFormer}
\centering
\begin{tabular}{ccccc}
\hline
\multirow{2}{*}{Methods} & {Focal Joints} & \multirow{2}{*}{Part Branch} & \multirow{2}{*}{JP-CA} & \multirow{2}{*}{Acc (\%)} \\
 & Selection &  &  & \\
\hline
Basic-SFormer & - & - & - & 87.8 \\
\hline
A   & $\surd$ & - & - & 88.3  \\
B   & $\surd$ & $\surd$  & - & 89.1 \\
C   & $\surd$ & $\surd$ & $\surd$ & \textbf{89.5} \\
\hline
\end{tabular}
\end{table}

As seen, model A selects the focal joints at stage 2 of the network and improves the performance of Basic-SFormer by $0.5\%$. This indicates that it is beneficial to identify discriminative joints. Then, model B introduces the part branch to network stage 2. This provides performance improvement of $0.8\%$ and reflects the spatial relations of intra-parts carry helpful global motion patterns. Finally, by adding the JP-CA into model C, the accuracy is further increased by $0.4\%$. This implies that the interactions between body parts and the selected focal joints are helpful for distinguishing actions.


{\bf Impact of Number of Focal Joints.} To explore the effect of selecting different number of focal joints, we test the models using different $K$ in FG-SFormer blocks at stage 2. Note that $K = 25$ means all the joints are used. As shown in Table~\ref{tab:topk}, the accuracy gradually improves as $K$ increases from 3 to 15, and then decreases when it further increases. This implies that the redundant or noisy joints indeed harm the recognition performance. In addition, too small number of focal joints are not enough to accurately identify the actions. 
\begin{table}[!htbp]
\caption{Comparison of classification accuracy using different number of focal joints.}
\label{tab:topk}
\centering
\tabcolsep=0.2cm
\begin{tabular}{ccccccccc}
\hline
\boldmath{$K$}  & 3  & 6 & 9 & 12 & 15 & 18 & 21 & 25   \\
\hline
\textbf{Acc} (\%)  & 88.6 & 88.9 & 89.0 & 89.2 & \textbf{89.5} & 89.2 & 89.3 & 89.1  \\
\hline
\end{tabular}
\end{table}

{\bf Effectiveness of FG-TFormer Block.} To evaluate the efficacy of FG-TFormer block, we build up experiments based on the complete network by modifying the FG-TFormer block only. The model using Basic-TFormer is taken as the baseline, which solely adopts the global MSA in temporal dimension. According to the results shown in Table \ref{tab:LGC-TFormer}, without the $\text{TCN}_{\text{dilate}}$ in MSA, the Basic-TFormer performs significantly worse than our FG-TFormer with a large margin of $-6.7\%$. 
Besides, by replacing Basic-TFormer with $\text{TCN}_{\text{dilate}}$, the performance is greatly improved by $6.3\%$. Finally, our FG-TFormer further achieves improvement of $0.4\%$ by integrating $\text{TCN}_{\text{dilate}}$ into self-attention mechanism. 
\begin{table}[!htbp]
\caption{Ablation study for components in FG-TFormer block. }
\label{tab:LGC-TFormer}
\centering
\begin{tabular}{cccc}
\hline
Methods & Global MSA  & $\text{TCN}_{\text{dilate}}$  & Acc (\%)  \\
\hline
Basic-TFormer    & $\surd$ & $\times$ & 82.8 \\
\hline
\multirow{2}{*}{FG-TFormer}  & $\times$ & $\surd$ & 89.1    \\
                              &$\surd$ &$\surd$ & \textbf{89.5}    \\
\hline
\end{tabular}
\end{table}

{\bf Configuration Exploration.} We explore different network configurations for stages 1 and 2 in our FG-STFormer by adjusting the number of layers $L_1$ and $L_2$. 
The total number of layers is fixed as $8$. The results are shown in Table~\ref{tab:net-layer}. Comparing models A, B and C, we can find that higher performance is obtained with more than $4$ layers used in stage 1, and the best performance is achieved by $L_1 = 6$ and $L_2 = 2$. The accuracy drops down when stage 2 is assigned less layers in model D. These observations indicate that it is necessary for stage 1 to sufficiently learn the relations among all the joints, otherwise the performance could be harmed by focusing on unreliable focal joints and part collaborations. 
\begin{table}[!htbp]
\caption{Comparison of different network configurations of our FG-STFormer.}
\label{tab:net-layer}
\centering
\begin{tabular}{cccc|cccc}
\hline
\multirow{2}{*}{Methods} & Stage 1 & Stage 2 & \multirow{2}{*}{Acc (\%)} &\multirow{2}{*}{Methods} & Stage 1 & Stage 2 & \multirow{2}{*}{Acc (\%)} \\
& $L_1$ & $L_2$ & & & $L_1$ & $L_2$ & \\
\hline
A & 4 & 4 & 88.8 & C & 6 & 2  & \textbf{89.5} \\
B & 5 & 3  & 89.0 &D & 7 & 1  & 89.0  \\
\hline
\end{tabular}
\end{table}

\begin{figure}[!htbp]
\centering
\includegraphics[scale=0.4]{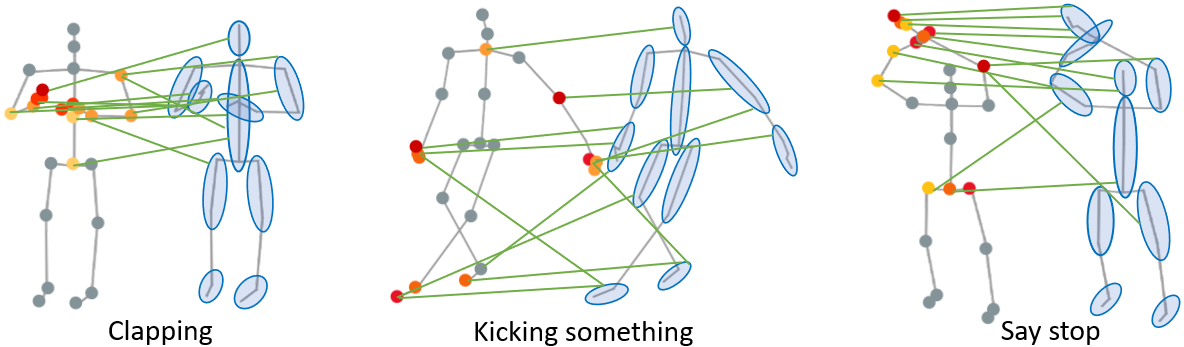}
\caption{The selected focal joints and learned joint-part interactions of actions.}
\label{fig:focal-joint}
\end{figure}
\subsection{Visualization and Analysis}
To validate what the focal joints are concentrated on at stage 2, we visualize the sampled 13 focal joints having largest scores for three actions in Fig.~\ref{fig:focal-joint}. These focal joints are depicted as coloured dots in the left skeleton of each action. The darker the dot is, the higher the informativeness score is of the joint.
We can see that the actions \textit{Clapping}, \textit{Kicking something} and \textit{Say stop} mainly select hands, shoulders, elbows and feet as the focal joints. Besides, Fig.~\ref{fig:focal-joint} illustrates the learned attention weights from parts to focal joints of these actions. Attentions with large values are shown as green lines. As seen, the actions \textit{Clapping} and \textit{Say stop} mainly build interactions between focal joints and upper limbs, while action \textit{Kicking something} interacts between focal joints and the whole body parts. These results verify that the spatial relations between the key joints and the global contextual movement information are captured by our FG-SFormer.

\begin{figure}[!htp]
\centering
\includegraphics[scale=0.38]{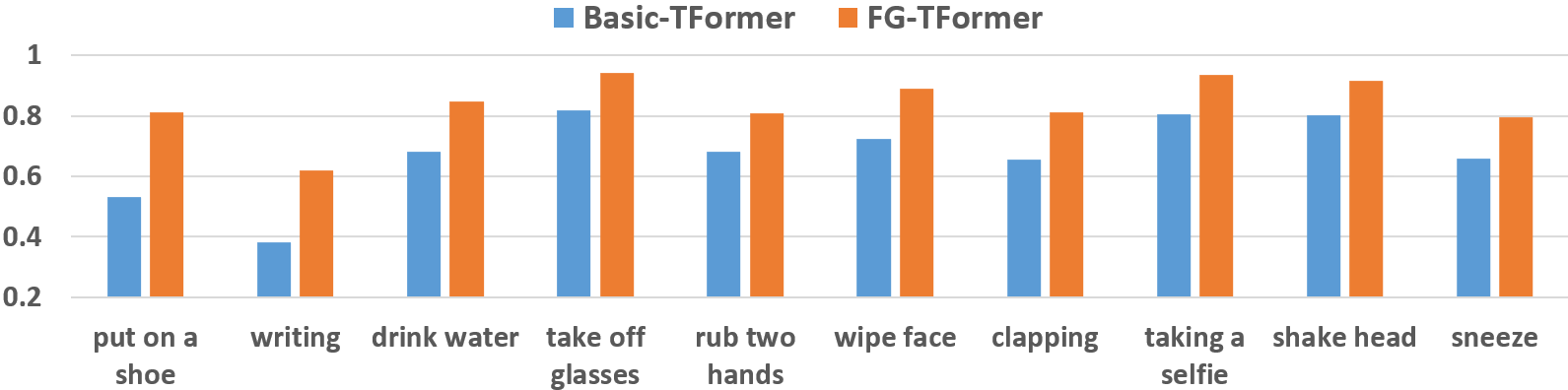}
\caption{Accuracy comparison between the Basic-TFormer and our FG-TFormer blocks.}
\label{fig:acc-compare}
\end{figure}

Moreover, we compare the performance of the Basic-TFormer with our FG-TFormer on action classes that the former has low accuracy. As shown in Fig.~\ref{fig:acc-compare}, our network improves the performance of those exhibited classes, which mainly involve the subtle and fine-grained motions of hands, feet and head. This concludes that our FG-TFormer can capture those subtle interaction patterns via explicitly embedding the neighboring relations into it.

\begin{table}[!htbp]
\caption{Comparison to state-of-the-arts on NW-UCLA dataset.}
\label{sota-ucla}
\centering
\begin{tabular}{lcc}
\hline
\multirow{2}{*}{Methods}  & \multirow{2}{*}{Year}  & NW-UCLA  \\
& & Top-1 (\%)\\
\hline
HBRNN-L~\cite{du2015hierarchical}  & 2015 & 78.5  \\
Ensemble TS-LSTM~\cite{lee2017ensemble} & 2017  & 89.2 \\
AGC-LSTM~\cite{si2019attention}  & 2019 & 93.3 \\
Shift-GCN~\cite{cheng2020skeleton}  & 2020 & 94.6  \\
DC-GCN+ADG~\cite{cheng2020decoupling} & 2020 & 95.3 \\
CTR-GCN~\cite{chen2021channel}   & 2021 & 96.5  \\
\hline
FG-STFormer (ours)  & 2022 & \textbf{97.0}   \\
\hline
\end{tabular}
\end{table}
\subsection{Comparison with the State-of-the-arts} 
We compare our FG-STFormer with existing state-of-the-art (SOTA) methods on three datasets: NW-UCLA, NTU-60 and NTU-120. Following the previous works~\cite{liu2020disentangling,zhang2021stst,shi2020decoupled}, we fuse results of four modalities, i.e., joint, bone, joint motion, and bone motion. The results are shown in Table~\ref{sota-ucla} and Table~\ref{tab:sota-ntu60-120}. As seen, our method outperforms all existing transformer-based methods under nearly all evaluation benchmarks on NTU-60 and NTU-120, including the latest method STST~\cite{zhang2021stst} which uses not only the parallel spatial and temporal transformers but also multiple self-supervised learning tasks, and ST-TR~\cite{plizzari2021skeleton} which adopts hybrid architecture of spatial-temporal transformer and GCN. Our method surpasses DSTA~\cite{shi2020decoupled} by $2.4\%$ and $1.6\%$ on the two evaluation protocols of NTU-120.
\begin{table}[!htbp]
\caption{Performance comparisons against the SOTA methods on NTU- 60 and 120.}
\label{tab:sota-ntu60-120}
\centering
\begin{tabular}{lccc|cc}
\hline
\multirow{2}{*}{{Methods}}
&\multirow{2}{*}{{Year}}
& \multicolumn{2}{c|}{{NTU-60}} &\multicolumn{2}{c}{{NTU-120}}\\
&&{X-Sub} (\%)  &{X-View} (\%) & {X-Sub} (\%)  &{X-Set} (\%)\\
\hline
\multicolumn{6}{l}{\textbf{GCN-based Methods}} \\
\hline
ST-GCN~\cite{yan2018spatial} & 2018	& 81.5  &  88.3 & 70.7  &  73.2   \\
2s-AGCN~\cite{shi2019two}	& 2019 & 88.5 & 95.1 & 82.9  & 84.9    \\
DGNN~\cite{shi2019skeleton}	& 2019	& 89.9  &96.1  &- &-   \\
Shift-GCN~\cite{cheng2020skeleton} & 2020	& 90.7  & 96.5 & 85.9  & 87.6    \\
Dynamic GCN~\cite{ye2020dynamic} & 2020	& 91.5  & 96.0  & 87.3  & 88.6   \\
MS-G3D~\cite{liu2020disentangling} & 2020	& 91.5  & 96.2   & 86.9  & 88.4    \\
MST-GCN~\cite{chen2021multi} & 2021	& 91.5  & 96.6  & 87.5  & 88.8 \\
CTR-GCN~\cite{chen2021channel} & 2021 & 92.4 & 96.8 & 88.9 & \textbf{90.6} \\
STF~\cite{ke2022towards} & 2022 & 92.5 & \textbf{96.9} & 88.9 & 90.0 \\
\hline
\multicolumn{6}{l}{\textbf{Transformer-based Methods}} \\
\hline
DSTA~\cite{shi2020decoupled} & 2020		& 91.5  & 96.4 & 86.6  & 89.0 \\
ST-TR~\cite{plizzari2021skeleton} & 2021	& 89.9  & 96.1 & 82.7  & 84.7     \\
UNIK~\cite{yang2021unik} & 2021	& 86.8  & 94.4  & 80.8  & 86.5   \\
STST~\cite{zhang2021stst} & 2021		& 91.9  & 96.8  &- &-   \\
\hline
FG-STFormer (ours)    & 2022 & \textbf{92.6}  & 96.7 & \textbf{89.0} & \textbf{90.6}\\
\hline
\end{tabular}
\end{table}

Moreover, compared to GCN-based methods, the performance of our FG-STFormer is also at the top. It compares favourably with current state-of-the-art STF~\cite{ke2022towards} and CTR-GCN~\cite{chen2021channel} on NTU-60 and NTU-120, and even outperforms the latter on NW-UCLA by $0.5\%$, verifying the effectiveness of FG-STFormer. 


\section{Conclusion}
In this work, we present a novel focal and global spatial-temporal transformer network (FG-STFormer) for skeleton-based action recognition. In spatial dimension, it learns intra- and inter- correlations for adaptively sampled focal joints and global body parts, which captures the discriminative and comprehensive spatial dependencies. In temporal dimension, it explicitly learns both the local and global temporal relations, enabling the network to capture rich motion patterns effectively. On three datasets, the proposed FG-STFormer achieves the state-of-the-art performance, demonstrating the effectiveness of our method.\\

%
%
%
%




\bibliographystyle{splncs04}
\bibliography{FG_STFormer_accv2022}

\end{document}